\documentclass[runningheads,table,xcdraw]{llncs}
\usepackage{graphicx}
\usepackage[table,xcdraw]{xcolor}
\usepackage{tikz}
\usepackage{comment}
\usepackage{amsmath,amssymb} 
\usepackage{color}
\usepackage{adjustbox} 
\usepackage[colorlinks, linkcolor=blue]{hyperref}
\usepackage[marginal]{footmisc}
\usepackage{orcidlink}
\usepackage{subfigure}
\usepackage{scrextend}

\usepackage[accsupp]{axessibility}  

\begin{document}
\pagestyle{headings}
\mainmatter
\def\ECCVSubNumber{5763}  

\title{3D Object Detection with a Self-supervised Lidar Scene Flow Backbone} 

\titlerunning{3D Object Detection with a Self-supervised Lidar Scene Flow Backbone}
%

\author{Ekim Yurtsever\inst{2 * }\orcidlink{0000-0002-3103-6052} \and
Emeç Erçelik\inst{1 * }\orcidlink{0000-0002-0716-0475} \and
Mingyu Liu\inst{1,3}\orcidlink{0000-0002-8752-7950}\and
Zhijie Yang\inst{1} \and
Hanzhen Zhang \inst{1} \and
Pınar Topçam \inst{1} \and
Maximilian Listl \inst{1} \and
Yılmaz Kaan Çaylı \inst{1} \and
Alois Knoll \inst{1}\orcidlink{0000-0003-4840-076X}}
\authorrunning{E. Yurtsever, E. Erçelik et al.}

%
\institute{Chair of Robotics, Artificial Intelligence and Real-time Systems,
        Technical University of Munich, 85748 Garching b. M\"{u}nchen, Germany \\
        \email{\{emec.ercelik, mingyu.liu, zhijie.yang, hanzhen.zhang, pinar.topcam, maximilian.listl, kaan.cayl \}@tum.de, knoll@in.tum.de} \\ \and
Ohio State University, Columbus, OH 43212, USA\\
\email{yurtsever.2@osu.edu}\\ \and
Tongji University, 201804, Shanghai, China \\ 
\footnote{* Authors contributed equally.}
}

\maketitle

\begin{abstract}
State-of-the-art lidar-based 3D object detection methods rely on supervised learning and large labeled datasets. However, annotating lidar data is resource-consuming, and depending only on supervised learning limits the applicability of trained models. Self-supervised training strategies can alleviate these issues by learning a general point cloud backbone model for downstream 3D vision tasks. Against this backdrop, we show the relationship between self-supervised multi-frame flow representations and single-frame 3D detection hypotheses. Our main contribution leverages learned flow and motion representations and combines a self-supervised backbone with a supervised 3D detection head. First, a self-supervised scene flow estimation model is trained with cycle consistency. Then, the point cloud encoder of this model is used as the backbone of a single-frame 3D object detection head model. This second 3D object detection model learns to utilize motion representations to distinguish dynamic objects exhibiting different movement patterns. Experiments on KITTI and nuScenes benchmarks show that the proposed self-supervised pre-training increases 3D detection performance significantly. \href{https://github.com/emecercelik/ssl-3d-detection.git}{https://github.com/emecercelik/ssl-3d-detection.git}


\end{abstract}

\section{Introduction}
Lidar promises accurate distance measurement, which is crucial for real-time systems such as 3D perception modules of automated vehicles. Supervised learning methods have dominated benchmarks created for challenging downstream 3D vision tasks. However, high-performing models need a copious amount of labeled data for training. Annotating lidar data is labor-intensive and is a bottleneck for real-world deployment. 

\begin{figure}[t]
  \centering
    \includegraphics[width=0.9\linewidth]{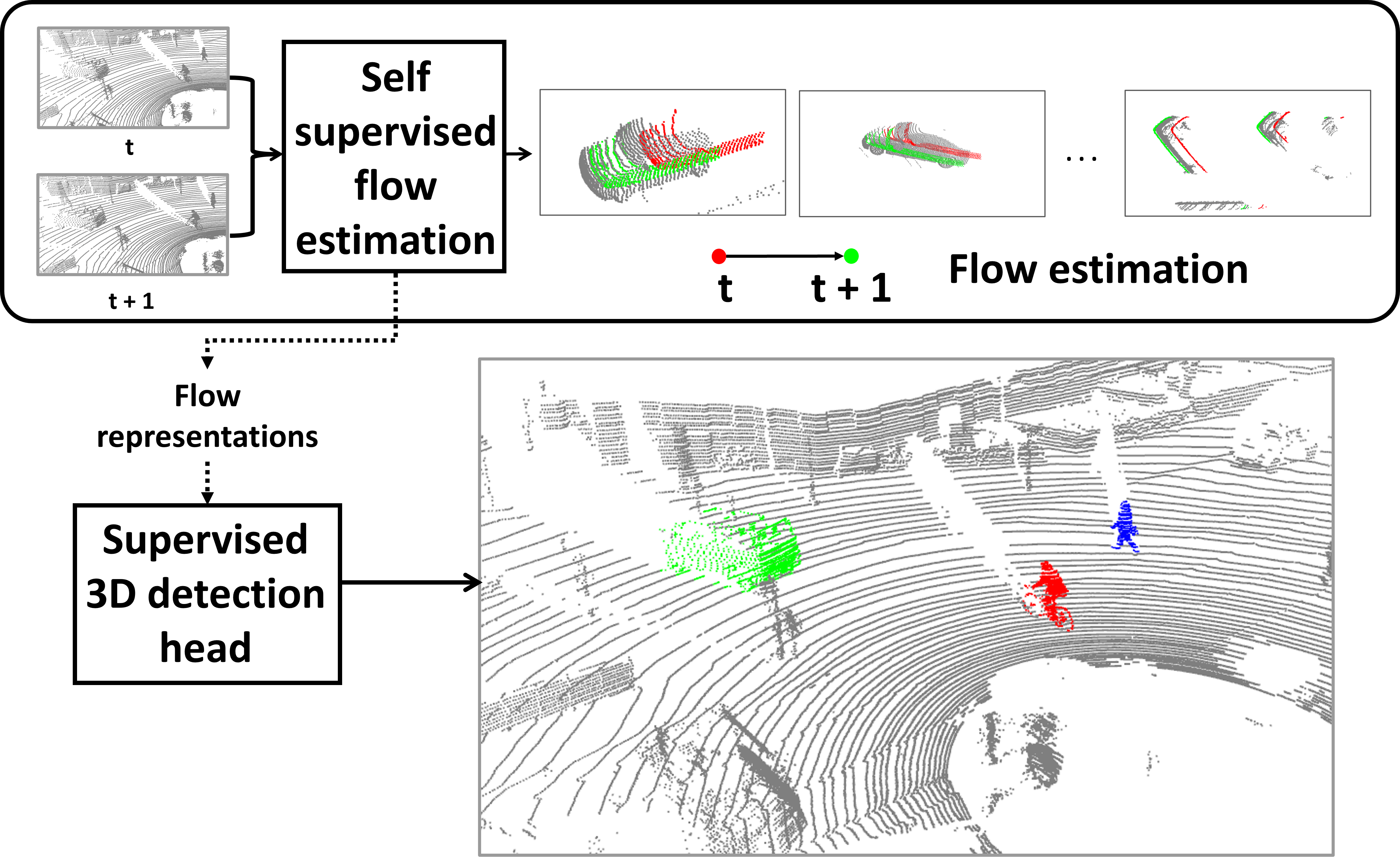}
   \caption{This study shows the relationship between self-supervised flow representations and supervised 3D detection hypotheses. We illustrate the importance of defining 3D objects-of-interest hypotheses in a spatio-temporal context. For example, a car should not just be defined by its shape but also by its capability to move in space and time. To this end, a scene flow estimation network is trained with cycle consistency in a self-supervised manner. Then, the backbone of this pre-trained model is used to feed a supervised 3D object detection head. The proposed strategy improves detection performance significantly compared to baselines when the amount of labeled data is less for supervised learning.}
   \label{fig:first_page}
\end{figure}

Recent work showed the importance of self-supervised learning to build large backbones by exploiting the structure of data. For example, the temporal contextual changes in videos can be exploited in contrastive learning strategies \cite{tschannen2020self}. Contrastive methods have also been used with data augmentation \cite{purushwalkam2020demystifying} for similar purposes. MoCo \cite{he2020momentum} classifies images in binary form as positive and negative to learn useful representations. Another approach is to quantize representations from a teacher network \cite{caron2020unsupervised}, \cite{asano2019self}. However, these works focus solely on the RGB image domain. Not much work focuses on unsupervised or self-supervised 3D object detection. Point cloud sparsity poses additional challenges, as the structure of data is significantly different from denser modalities.

The main body of state-of-the-art 3D object detection with point cloud literature comprises supervised learning methods \cite{zhou2018voxelnet,yan2018second,he2020structure,shi2020pv,yang20203dssd,shi2019pointrcnn,yang2019std,shi2020point,deng2021voxel,lang2019pointpillars,ye2020hvnet}. 

Point cloud scene flow is another important 3D vision task. Initially, supervised learning was shown to be superior for the task \cite{wang2021hierarchical,liu2019meteornet,gu2019hplflownet,wei2021pv,liu2019flownet3d,wang2020flownet3d++,puy2020flot}. More recently, self-supervised and unsupervised 3D scene flow and stereo flow methods have been introduced \cite{wang2015unsupervised,basha2013multi,hur2020self,yin2018geonet,zou2018df,lai2019bridging,chen2019self,godard2019digging,ranjan2019competitive,wang2019learning,luo2019every}. However, these developments have not been utilized for the 3D object detection task up until now.


We propose to employ a scene flow backbone trained in a self-supervised fashion to learn useful representations that other downstream head models can utilize after fine-tuning, such as a 3D object detection head (Fig. \ref{fig:first_page}). First, we follow the cycle-consistency approach \cite{mittal2020just} to train a FlowNet3D-based \cite{liu2019flownet3d} scene-flow backbone using self-supervised learning. We introduce architectural changes to the FlowNet3D module to incorporate a point cloud backbone that can also be utilized with a detection head. We explore several training and loss strategies, including auxiliary training, to find the best layout. Empirical evidence obtained from KITTI \cite{geiger2012we} and nuScenes \cite{caesar2020nuscenes} datasets show that the proposed strategy increases 3D detection performance significantly.

Our method differs from \cite{mittal2020just} in two important ways. To the best of our knowledge, lidar point cloud-based 3D object detection has not been successfully achieved with a self-supervised backbone up until now. Our modifications to the FlowNet3D \cite{liu2019flownet3d} architecture enables the integration of point-level temporal changes with 3D detection. Secondly, our combined auxiliary training cycle consistency and supervised 3D detection losses lead to learning more general representations as well as motion representations, which identify objects based on their contextual motion patterns.

A summary of our main contributions is as follows:

\begin{itemize}
    \item Employing self-supervised point cloud scene flow estimation to learn motion representations for 3D object detection in tandem with supervised fine-tuning
    \item We show that auxiliary training is the best strategy for using self-supervised cycle-consistency loss along with supervised 3D detection loss.
    \item The proposed strategy is especially effective with a lesser amount of supervised data. We obtained a significant performance boost when only a smaller part of supervised training data was used for the 3D detection task.
\end{itemize}

\section{Related Work}

\textbf{Scene flow.} Scene flow was first introduced as an extension of optical flow in the third dimension and was estimated with a linear computation algorithm\cite{vedula2005three}. Stereo cameras \cite{huguet2007variational,vogel20153d} and RGB-D were also \cite{shao2018motion} utilized to derive scene flow. Current state-of-the-art uses lidar point clouds and deep neural networks to estimate scene flow with supervised learning \cite{liu2019flownet3d,wang2020flownet3d++,gu2019hplflownet}. Most commonly, two subsequent lidar frames are used to estimate the flow vectors of each point in the scene. Building ground truth for such vectors is labor-intensive. As such, synthetic datasets are more popular for scene flow benchmarking \cite{puy2020flot}. 

\textbf{Self-supervised scene flow estimation.} Self-supervised scene flow estimation is a relatively understudied angle. A recently proposed solution \cite{mittal2020just} is to use cycle-consistency and nearest neighbors losses to train an estimation network.  Several other distance metrics and regularization techniques such as using chamfer distance, smoothing, and regularization \cite{wu2019pointpwc} have also been employed for the same task. A more recent study showed that self-supervised scene flow could also be combined with motion segmentation \cite{baur2021slim}.

\textbf{Self-supervised 3D Object Detection.} A monocular 3D object detection model has been trained with self-supervised learning using shape priors and 2D instance masks \cite{beker2020monocular}. Another monocular 3D object detection model with weak supervision has been trained using shape priors \cite{wang2020directshape}. \cite{rao2021randomrooms} generates random synthetic point cloud scenes for pre-training to learn useful representations from CAD models. \cite{xie2020pointcontrast,liang2021exploring,hou2021exploring} mainly use contrastive pre-training to learn geometrical point cloud representations with different views of the same scene.  However, there is not much work focusing on self-supervised 3D object detection considering motion representations with point clouds. We aim to fill this gap in the literature.

\section{Method}
\begin{figure}
  \centering
   \includegraphics[width=1\linewidth]{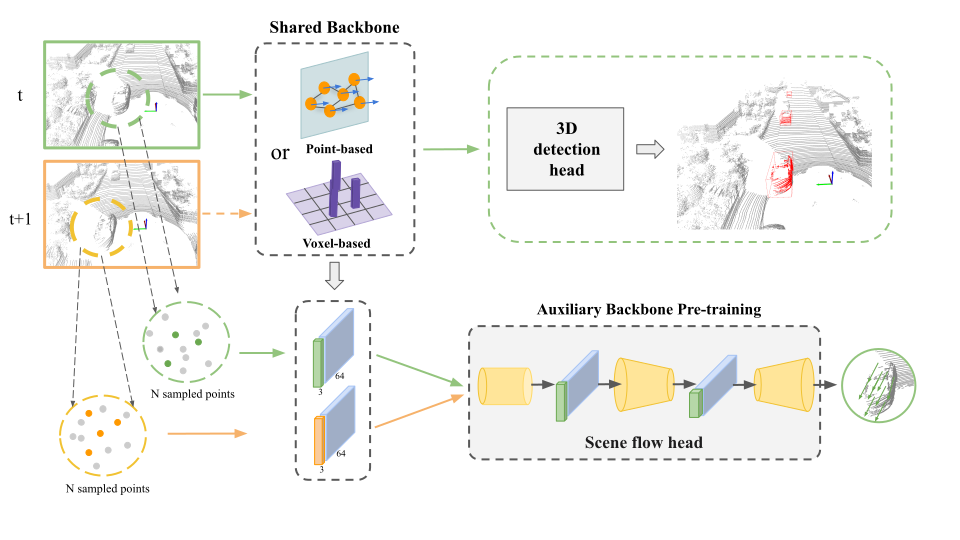}
   \caption{Our self-supervised 3D object detection pre-training: The auxiliary scene flow head is used to train the 3D detection backbone (point- or voxel-based) for motion-aware point cloud representations with self-supervised cycle consistency loss \cite{mittal2020just}. The motion representations learned without labelled data can help distinguish objects based on their motion patterns for a 3D downstream task. Then, we further train the pre-trained backbone and a 3D detection head for 3D detection with labelled data.}
   \label{fig:overview}
\end{figure}

Backbone of a 3D object detector is mainly used to extract point, voxel, or region features to detect possible objects in that vicinity. Due to the limitation in labelled dataset sizes, we aim to train the backbone on a large unlabelled dataset using self-supervision to obtain good motion-aware point representations. Afterwards, it is possible to use the pre-trained backbone for the 3D detection supervised training with a smaller dataset. Thus, the 3D detection network can benefit from the initialized point motion representations to distinguish objects based on movement patterns. We summarize our method in Fig. \ref{fig:overview}.

\subsection{Problem Definition}

Given two subsequent lidar point cloud frames $\mathcal{P}_t = \{\textbf{p}_i\}_M, \textbf{p}  \in 	\mathbb{R}^{3}$ and $\mathcal{P}_{t+1} \in \mathbb{R}^{N \times 3}$, we are first interested in estimating the scene flow $\mathcal{F}_{t \rightarrow t+1} = \{d_i\}_M $, where $\textbf{d}_i = \textbf{p}_i' - \textbf{p}_i$. $\textbf{p}_i'$ denotes the new position at time $t + 1$ of point $i$ in the first point cloud $\mathcal{P}_t$. It should be noted that the second point cloud may or may not contain a point corresponding to $\textbf{p}_i'$ due to sparsity. The second objective is to map $\mathcal{P} \rightarrow \{T_j\}_U$, where $T_j$ is the 3D object detection tuple containing class id and bounding box shape and coordinates, using previously-learned spatio-temporal representations. $U$ is the total number of objects in the point cloud frame. 

We aim to benefit from the point motion representations learned by the 3D feature extractor, $g$, during self-supervised scene flow training $\mathcal{F}_{t \rightarrow t+1} = s(g(\mathcal{P}_t,\mathcal{P}_{t+1}))$, where $s$ is the scene flow head. In this way, the 3D detection head, $h$, can use the spatio-temporal motion representations learned in $g$ to better identify complex object point patterns for meaningful detection results such that $\{T_j\}_U = h(g(\mathcal{P}))$. 

\subsection{Self-supervised Scene Flow}
\textbf{Backbone:} We first follow the cycle-consistency approach \cite{mittal2020just} to train a scene flow estimator. We use a 3D detector's backbone to extract local features of sampled points from two consecutive frames. This allows the self-supervised scene flow gradients to be backpropagated through the backbone. Hence, the backbone learns point representations encoding object movement patterns. The learned spatio-temporal features can be further used to distinguish objects from the background and other objects for the 3D perception task. 


\textbf{Scene Flow Head:} The scene flow head based on FlowNet3D \cite{liu2019flownet3d} generates flow embeddings from local point features provided by the 3D backbone. The shape of input points, $\mathcal{P}_t$, is reconstructed by applying \textit{set upconv layers} to local flow embeddings for the final scene flow estimations  $\mathcal{F}_{t \rightarrow t+1}$. 

\textbf{Training with Cycle Consistency:} We use 3D detection backbone as the feature extractor for the scene flow head. Both the backbone and the scene flow head are trained with the self-supervised cycle consistency loss given in \cite{mittal2020just}. For the cycle consistency, the scene flow is calculated in forward and backward directions, meaning $\mathcal{F}_{t \rightarrow t+1}$ and $\mathcal{F}_{t+1 \rightarrow t}$. The $\mathcal{F}_{t+1 \rightarrow t}$ makes use of the new positions of the propagated points $\textbf{p}_i'$ to close the cycle. The $\textbf{p}_i''$ is the estimated positions of the $\textbf{p}_i$ in the backward direction with the $\mathcal{F}_{t+1 \rightarrow t}$. The mismatch between the $\textbf{p}_i$ and the $\textbf{p}_i''$ at frame $t$ allows training of the backbone in a self-supervised way. With the self-supervised training, 3D backbone learns to generate regional flow and motion features from the given set of point clouds. 

\subsection{Downstream Task: 3D Object Detection}
We are interested in the 3D object detection as the 3D downstream task. The scene flow head, $s$, and the 3D object detection head, $h$, use the same backbone, $g$, as seen in Fig. \ref{fig:overview}. Also, point- or voxel-based 3D backbone encodings can be used. We initialize 3D detector's backbone weights with the pre-trained weights from the auxiliary self-supervised scene flow training. In this way, we assume that the pre-trained backbone from scene flow can already provide good geometry- and motion-aware point features. The 3D detection head takes distinguishable spatio-temporal point cloud features based on different object motion patterns. Hence, the 3D detection network can detect objects more accurately even after supervised training with a smaller labelled dataset. We show the efficacy of our approach in section \ref{sec:results}. Note that the scene flow head is for the auxiliary scene flow training and is not used during the 3D detection training and inference. 

\section{Implementation Details}
\label{sec:implementation}
Our self-supervised auxiliary backbone pre-training approach can be used with different 3D detector architectures. We evaluate our method with mainly three different 3D detectors, Point-GNN\footnote{\href{https://github.com/WeijingShi/Point-GNN}{https://github.com/WeijingShi/Point-GNN}}\cite{shi2020point}, CenterPoint\footnote{\label{footnote_mmdet}\href{https://github.com/open-mmlab/mmdetection3d/}{https://github.com/open-mmlab/mmdetection3d/}}\cite{yin2021center}, and PointPillars\footref{footnote_mmdet}\cite{lang2019pointpillars}, which are point-, voxel-based approaches. For the self-supervised scene flow task, we add the modified FlowNet3D as well as the cycle-consistency loss\footnote{\href{https://github.com/HimangiM/Just-Go-with-the-Flow-Self-Supervised-Scene-Flow-Estimation}{https://github.com/HimangiM/Just-Go-with-the-Flow-Self-Supervised-Scene-Flow-Estimation}} \cite{mittal2020just} to 3D detectors' training pipelines.


\subsection{Pre-training with Self-supervised Scene Flow}
The FlowNet3D takes a set of points from two successive frames as input and estimates the flow vectors. The network extracts the point features with two cascaded \textit{PointNet Set Abstraction} modules, each with a 3-layer MLP. We remove the first \textit{PointNet Set Abstraction} module and feed in the point features from 3D detector's backbone to the second \textit{PointNet Set Abstraction} module. 

\textbf{Point cloud backbone:} We use Point-GNN, PointPillars, and CenterPoint backbones for our main and ablation results. The CenterPoint and PointPillars backbones have the same architecture as we use mmdetection3d \cite{mmdet3d2020} implementations. Point-GNN extracts keypoint features from a 3-level graph network used as a backbone, from which we obtain keypoint features of two consecutive point clouds. After sampling $N$ points from each frame, we apply bilinear interpolation to get features of the sampled points from keypoint features according to their positions. We use the settings provided for the best performing Point-GNN with $T=3$, which represents the number of graph levels. For the CenterPoint and PointPillars detectors, we follow a similar approach and use their voxel encoders to obtain features of the sampled points from two consecutive frames without making any changes to the 3D detector's architecture. 

\textbf{Scene flow head:} Scene flow head is responsible for estimating 3D motion of the points between two sequential frames. The FlowNet3D's scene flow head consists of \textit{flow embedding}, \textit{set conv layers}, and \textit{set upconv layers} followed by fully-connected layers for estimating point flow vectors. The scene flow head takes the local point features as inputs. We remove only the final \textit{set upconv layer} that takes skip connections from the first \textit{PointNet Set Abstraction} module, which we replace with the 3D detector's backbone. 


\textbf{Training strategy:} We train the point cloud backbone and scene flow head end-to-end using the self-supervised scene flow loss. For the scene flow training on Point-GNN backbone, we initialize our scene flow head weights with the pre-trained FlowNet3D weights on the supervised FlyingThings3D \cite{mayer2016large} simulation data, we use Stochastic Gradient Descent (SGD) optimizer with $6.25\times 10^{-5}$ learning rate. The number of sampled points from each frame is $N=2048$. We train the scene flow network for 80k steps on the KITTI tracking dataset without using any labels. The model is trained on a single Nvidia Tesla V100 GPU. We train the PointPillars scene flow network on KITTI tracking dataset with $0.01$ learning rate. We sample $2048$ points per frame to train the model with batch size $2$ on one Nvidia RTX 2080 GPU. For the PointPillars- and CenterPoint-based scene flow training on nuScenes dataset, we use Adam optimizer using the voxel encoders as the bakcbone with a $0.001$ learning rate. Our batch size is $2$ for $N=2048$ sampled points. We train the network for $4$ epochs on one Nvidia RTX 2080 GPU.

\subsection{3D Detection Fine-tuning}

\textbf{3D detection heads:} We use Point-GNN, CenterPoint, and PointPillars as the 3D detectors for our results to show the efficacy of our self-supervised scene flow pre-training approach. We initialize detectors' backbone weights with the pre-trained backbone weights from the auxiliary scene flow task for a better point feature representation. 

\textbf{Training strategy:} After initializing weights of the 3D detector backbone from the scene flow task, we further train the backbone and the 3D detection heads with the 3D detection loss. We apply an alternating training strategy between the self-supervised scene flow and supervised 3D detection trainings: \textbf{(i)} Train the backbone and the scene flow head for self-supervised scene flow, \textbf{(ii)} train the pre-trained backbone and the detection head with 3D detection training, \textbf{(iii)} train the backbone from step (ii) and the scene flow head from step (i) for the scene flow, and finally \textbf{(iv)} train the backbone from step (iii) and the detection head from step (ii) for 3D detection.

We train the Point-GNN baseline for $1400k$ steps using the SGD optimizer with a learning rate of $0.125$ as done in the Point-GNN paper. For the training in step \textbf{(ii)} and step \textbf{(iv)}, we use SGD with a learning rate of $0.1$. The trainings took place on an Nvidia Tesla V100 GPU. We use batch size $4$ for all Point-GNN trainings. The PointPillars detector is trained for $24$ epochs with a learning rate of $0.001$ using AdamW optimizer. We used two Nvidia RTX 2080 Ti GPUs for the training. We use batch size $2$ for the PointPillars scene flow training and batch size $4$ for the detection training. For the training of CenterPoint detector, we keep the default setting in mmdetection3d, which is trained for 20 epochs with a learning rate of 0.001 using the AdamW optimizer. We set the batch size $2$ for the scene flow traing and batch size $20$ for detection head training. We use one Nvidia RTX 3090 GPU for the training. 

\subsection{Datasets}
We use KITTI 3D Object Detection, KITTI Multi-object Tracking datasets \cite{geiger2012we} as well as nuScenes dataset \cite{caesar2020nuscenes} for the self-supervised scene flow and the supervised 3D detection training and validation. 

\textbf{KITTI 3D Object Detection:} KITTI 3D Object Detection dataset consists of 7481 training frames sampled from different drives. Since the provided frames are not sequential, we use lidar point clouds only for 3D object detection training. Only objects visible in the camera-view are annotated. We utilize the common train-val split with $3712$ training and $3769$ validation samples. For the evaluation, the KITTI average precision (AP) metric is used for three different difficulty levels with $IoU=0.7$ for the car class.

\textbf{KITTI Multi-object Tracking:} This dataset contains 21 training and 29 testing drives, each of which consists of several sequential frames. We use the tracking dataset only for the self-supervised scene flow training without using any annotations. Therefore, we combine all the training and testing drive data for training except the training drives $11$, $15$, $16$, and $18$, which are used for observing cycle consistency validation loss. This gives us $11902$ frames for self-supervised scene flow training.

\textbf{nuScenes: } nuScenes is also an autonomous driving dataset, which consists of 700 training and 150 validation drives. The annotations are provided at 2 Hz for 360-degree objects and the lidar sweeps are collected at 20 Hz. nuScenes is a larger dataset than KITTI and it is collected from denser and more challenging environments. There are 10 classes annotated in the nuScenes dataset. The main metrics are the average precision (AP) per class, mean average precision (mAP) among all classes, and the nuScenes detection score (NDS). Since the provided data contains sequential lidar point clouds, we use this dataset for both self-supervised scene flow and 3D detection trainings. 

\subsection{Loss}

For the 3D object detection training, we keep the same loss functions used for the 3D detectors.

\textbf{Point-GNN\cite{shi2020point}:} Point-GNN combines localization, classification, and regularization losses. Classification loss is calculated with the average cross-entropy loss among four classes, which are \textit{background}, horizontal and vertical anchor box classes, and a \textit{don't care} class. The network regresses $7$ bounding box parameters, $(x,y,z)$ for the center coordinates, $(w,h,l)$ for the width, height, and length sizes, and $\theta$ for the orientation of the bounding box. The Huber loss and $L1$ regularization are used as regression and regularization losses, respectively. We use the original loss coefficients for the total loss.

\textbf{PointPillars\cite{lang2019pointpillars}:} Similarly, PointPillars regresses the $7$ bounding box parameters and utilizes the Huber loss as a regression loss. The orientation is predicted from a set of discrete bins, for which the softmax classification loss is utilized. The focal loss is used as a classification loss for the object classes. All the loss values are combined with respective coefficients for the total loss and we use the default values given in the \textit{mmdetection3d} \cite{mmdet3d2020} repository. 

\textbf{CenterPoint\cite{yin2021center}:} CenterPoint is trained with the focal loss for the heatmap-based classification and the binary cross entropy loss for the IoU-based confidence score. Huber loss is used for the regression of box parameters. We keep the default values given in the \textit{mmdetection3d} \cite{mmdet3d2020} repository.


\textbf{Self-supervised scene flow loss:} We utilize the self-supervised loss used in \cite{mittal2020just} for training the 3D detector backbone and the scene flow head. The loss consists of the nearest neighbor and the cycle consistency losses. The nearest neighbor loss calculates the Euclidean distance of the point $\textbf{p}_i'$ to its nearest neighbor in frame $t+1$. $\textbf{p}_i'$ is the point transformed from frame $t$ to $t+1$. For the cycle consistency loss calculation the flow is applied in the forward ($\mathcal{F}_{t \rightarrow t+1}$) and the backward ($\mathcal{F}_{t+1 \rightarrow t}$) directions. The distance between the resulting $\textbf{p}_i''$ point and its anchor $\textbf{p}_i$ is used for the cycle consistency loss. Both losses are summed up for the total loss and only this loss is used for the training of the backbone and the scene flow head. 

\subsection{Experiments}

We conduct several experiments to show the efficacy of our self-supervised pre-training method on 3D object detection task using Point-GNN, PointPillars, CenterPoint, and SSN \cite{zhu2020ssn} 3D detectors. First, we compare our self-supervised pre-trained Point-GNN, CenterPoint, and PointPillars with their baselines trained with $100\%$ of the annotated 3D detection data. We show our results on KITTI and nuScenes validation and test sets. Then, we check the performance of our self-supervised detectors and their baselines in the low-data regime. For this, we apply supervised training using only using a smaller part of the annotated data in the ablation study. We also report detection accuracy of self-supervised detectors trained with and without alternating training strategy to justify our alternating self-supervised scene flow and supervised 3D detection scheme. Finally, we compare our self-supervised scene flow pre-training method against other self-supervised learning methods.


\begin{figure}[h]
    \centering
    \includegraphics[width=0.95\textwidth]{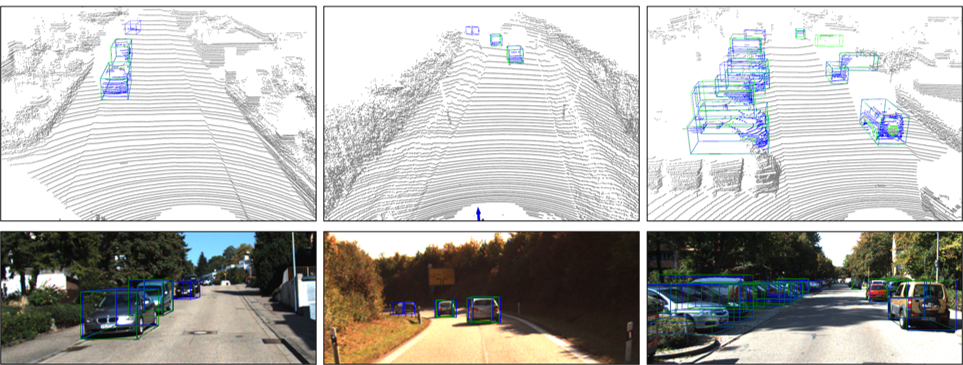}
    \caption{Qualitative Results: Three different scenes from the KITTI 3D Detection validation set in the columns. Blue and green bounding boxes are for our approach and the baseline Point-GNN \cite{shi2020point}, respectively. Our method can detect a distant hard object (left-most column) and a moving distant car (middle column) while the baseline misses it. Our method also performs better in a denser environment (right-most column).}
    \label{fig:pointgnn_qualitative}
\end{figure}


\section{Results \& Discussion}
\label{sec:results}

In this section, we provide our main results obtained using our scene flow-based self-supervised training with Point-GNN, CenterPoint, and PointPillars 3D detectors on KITTI and nuScenes validation and test sets. 



\begin{table}[h]
  \centering
\begin{tabular}{ccccccc}
 Car (IoU=0.7)  & \multicolumn{3}{c}{3D AP }&\multicolumn{3}{c}{BEV AP } \\ \hline
\multicolumn{1}{c|}{Method}&Easy&Mod& \multicolumn{1}{c|}{Hard}&Easy& Mod  & Hard  \\ \hline
\multicolumn{1}{c|}{Point-GNN* \cite{shi2020point}} & 90.44 & 82.12 & \multicolumn{1}{c|}{77.70} & 93.03 & 89.31 & 86.86 \\
\multicolumn{1}{c|}{\textbf{Self-supervised Point-GNN}}&\textbf{91.43}&\textbf{82.85} & \multicolumn{1}{c|}{\textbf{80.12}} & \textbf{93.55} & \textbf{89.79}  & \textbf{87.23}   \\

\multicolumn{1}{c|}{Improvement} & \multicolumn{1}{c}{{\color[HTML]{009901} +0.99}} & \multicolumn{1}{c}{{\color[HTML]{009901} +0.73}} & \multicolumn{1}{c}{{\color[HTML]{009901} +2.42}} & \multicolumn{1}{c}{{\color[HTML]{009901} +0.52}} & \multicolumn{1}{c}{{\color[HTML]{009901} +0.48}} & \multicolumn{1}{c}{{\color[HTML]{009901} +0.37}} \\ \hline
\multicolumn{1}{c|}{PointPillars} & 85.41 & 73.98 & \multicolumn{1}{c|}{67.76} & 89.93 & 86.57 & 85.20 \\
\multicolumn{1}{c|}{\textbf{Self-supervised PointPillars}} & \textbf{85.92} & \textbf{76.33} & \multicolumn{1}{c|}{\textbf{74.32}} & \textbf{89.96} & \textbf{87.44} & \textbf{85.53} \\

\multicolumn{1}{c|}{Improvement} & {\color[HTML]{009901} +0.51} & {\color[HTML]{009901} +2.36} & \multicolumn{1}{c|}{{\color[HTML]{009901} +6.56}} & {\color[HTML]{009901} +0.03} & {\color[HTML]{009901} +0.87} & {\color[HTML]{009901} +0.33} \\ \hline

\end{tabular}
  \caption{Self-supervised Point-GNN $\&$ PointPillars compared with the baseline on KITTI val. set for car class using 3D AP$_{R_{40}}$ metric. (*Reproduced baseline results for AP$_{R_{40}}$.)}
  \label{tab:main_kitti_40}
\end{table}

\begin{table}[h]
  \centering
\begin{tabular}{ccccccc}
 Car (IoU=0.7)  & \multicolumn{3}{c}{3D AP }&\multicolumn{3}{c}{BEV AP } \\ \hline
\multicolumn{1}{c|}{Method}&Easy&Mod& \multicolumn{1}{c|}{Hard}&Easy& Mod  & Hard  \\ \hline
\multicolumn{1}{c|}{AVOD\cite{ku2018joint}} & 76.39 & 66.47 & \multicolumn{1}{c|}{60.23}  & 89.75 & 84.95 & 78.32 \\
\multicolumn{1}{c|}{F-PointNet\cite{qi2018frustum}} & 82.19 & 69.79 & \multicolumn{1}{c|}{60.59}  & 91.17 & 84.67 & 74.77\\
\multicolumn{1}{c|}{TANet\cite{liu2020tanet}} & 84.39 & 75.94 & \multicolumn{1}{c|}{68.82}  & 91.58 & 86.54 & 81.19  \\
\multicolumn{1}{c|}{Associate-3Ddet\cite{du2020associate}} & 85.99 & 77.40 & \multicolumn{1}{c|}{70.53}  & 91.40 & 88.09 & 82.96 \\
\multicolumn{1}{c|}{UBER-ATG-MMF\cite{liang2019multi}} & 88.40 & 77.43 & \multicolumn{1}{c|}{70.22}  & 93.67 & 88.21 & 81.99 \\
\multicolumn{1}{c|}{CenterNet3D\cite{wang2020centernet3d}} & 86.20 & 77.90 & \multicolumn{1}{c|}{73.03}  & 91.80 & 88.46 & 83.62 \\
\multicolumn{1}{c|}{SECOND\cite{yan2018second}} & 87.44 & 79.46 & \multicolumn{1}{c|}{73.97}  & 92.01 & 88.98 & 83.67 \\
\multicolumn{1}{c|}{SERCNN\cite{zhou2020joint}} & 87.74 & 78.96  & \multicolumn{1}{c|}{74.30}  & 94.11 & 88.10 & 83.43 \\ \hline


\multicolumn{1}{c|}{PointPillars} & 80.51 & 68.57 & \multicolumn{1}{c|}{61.79} & \textbf{90.74} & 84.98 & 79.63 \\
\multicolumn{1}{c|}{\textbf{Self-supervised PointPillars}} & \textbf{82.54} & \textbf{72.99} & \multicolumn{1}{c|}{\textbf{67.54}} & 88.92 & \textbf{85.73} & \textbf{80.33} \\

\multicolumn{1}{c|}{Improvement} & {\color[HTML]{009901} +2.03} & {\color[HTML]{009901} +4.42} & \multicolumn{1}{c|}{{\color[HTML]{009901} +5.75}} & {\color[HTML]{FE0000} -1.82} & {\color[HTML]{009901} +0.75} & {\color[HTML]{009901} +0.7} \\ \hline

\end{tabular}
  \caption{Self-supervised PointPillars compared with the baseline on KITTI test set for car class using 3D AP$_{R_{40}}$ metric. }
  \label{tab:kitti_40_test}
\end{table}

\vspace{-1.2cm}

\subsection{Point-GNN}
The self-supervised Point-GNN is pre-trained on the scene flow task with cycle consistency loss using KITTI Tracking dataset without any annotations. Following, it is trained with the annotated KITTI 3D detection dataset using the proposed alternating training scheme. The baseline is trained using the same configuration and hyper-parameters. The only difference is that the baseline network weights are initialized randomly. Table \ref{tab:main_kitti_40} shows the 3D and BEV AP$_{R_{40}}$ scores of the baseline and self-supervised Point-GNN on KITTI validation set, where our method outperforms the baseline in all difficulty levels and especially with a large margin in hard difficulty level ($~2.5\%$). Similarly, our self-supervised Point-GNN outperforms its baseline on the KITTI test set on hard difficulty level with a $~2\%$ improvement. In supplementary material, we also include the same comparison with AP$_{R_{11}}$ metric using the reported Point-GNN results, where our method outperforms the original Point-GNN. These results show that motion-related point representations help distinguish even difficult objects that reflect only a small number of points. 

The Fig. \ref{fig:pointgnn_qualitative} shows our qualitative results on KITTI 3D Object Detection scenes.  The blue bounding boxes and green bounding boxes indicate results of our approach and the baseline, respectively. We show the bird's eye view and front-view lidar visualizations at the top and middle rows. At the bottom, we show the projected 3D bounding boxes on the image plane. Our approach can detect distant objects (left-most column) better as well as distant and moving objects (middle column). In addition, as seen in the right-most column of Fig. \ref{fig:pointgnn_qualitative}, our approach can provide better detection results in a denser scene. 





\subsection{CenterPoint}
Our self-supervised CenterPoint also outperforms the CenterPoint baseline on nuScenes validation and test sets as the mAP and NDS results given in Tables \ref{tab:nuscenes_validation} and \ref{tab:nuscenes_test}, respectively. We obtained the baseline scores with the best-performing \textit{mmdetection3d} CenterPoint checkpoint\cite{mmdet3d2020} for both evaluation sets. 


\subsection{PointPillars}



\begin{table}
  \centering
\begin{tabular}{c|cc|cccccccc}
Method & mAP & NDS & Car & Ped & Bus & Barrier & T. C. & Truck & Trailer & Moto.  \\ 
\hline
SECOND\cite{yan2018second} & 27.12 & - & 75.53 & 59.86 & 29.04 & 32.21 & 22.49 & 21.88 & 12.96 & 16.89 \\ 
PointPillars*\cite{lang2019pointpillars}& 40.02 & 53.29 & 80.60 & 72.40 & 46.30 & 52.60 & 33.60 & 35.10 & 26.20 & 38.40 \\
\begin{tabular}[c]{@{}c@{}}\textbf{Self-supervised} \\ \textbf{PointPillars}\end{tabular} & \textbf{42.06} & \textbf{55.02} & \textbf{81.10} & \textbf{74.50} & \textbf{49.50} & \textbf{54.70} & \textbf{34.70} & \textbf{38.40} & \textbf{29.70} & \textbf{38.80} \\ \hline
CenterPoint*\cite{yin2021center}& 49.13 & 59.73 & 83.70 & 77.40 & \textbf{61.90} & 59.40 & \textbf{52.90} & 50.20 & 35.00 & \textbf{44.40} \\ 

\begin{tabular}[c]{@{}c@{}}\textbf{Self-supervised} \\ \textbf{CenterPoint}\end{tabular} &\textbf{49.94} & \textbf{60.06} & \textbf{84.10} & \textbf{77.90} & 61.50 & \textbf{61.00} & 52.50 & \textbf{51.00} & \textbf{35.20} & 44.10\\ \hline

\end{tabular}
  \caption{Self-supervised PointPillars results on nuScenes validation set.(* mmdetection3d PointPillars checkpoint results, on which we built our work.) }
  \label{tab:nuscenes_validation}
\end{table}


We also report results of our self-supervised pre-training method using PointPillars \cite{lang2019pointpillars} 3D detector on the nuScenes and KITTI datasets. We pre-train the PointPillars voxel encoder with the self-supervised scene flow task without annotations. Following, the entire PointPillars network is trained on the annotated 3D detection data using our alternating training strategy. We compare our self-supervised PointPillars with its baseline on the KITTI validation and test sets as given in Tables \ref{tab:main_kitti_40} and \ref{tab:kitti_40_test}, respectively. Consistent with the previously-introduced results, our method improves the baseline results with a large margin for the 3D detection task. The increment is the most obvious for 3D AP moderate and hard difficulty levels with $~2.4\%$ and $~6.6\%$ for the validation set and with $~4.4\%$ and $~5.8\%$ for the test set.

In Table \ref{tab:nuscenes_validation}, we compare our self-supervised PointPillars with the baseline on nuScenes validation set. The baseline results are obtained from the best checkpoint given in the well-known mmdetection3d repository \cite{mmdet3d2020}. Our self-supervised PointPillars outperforms the baseline with a large increment on mAP and NDS metrics ($~2\%$) as well as for all class scores. Moreover, we provide results of our self-supervised PointPillars on nuScenes test set in Table \ref{tab:nuscenes_test} comparing to the previously-submitted PointPillars versions from the nuScenes leaderboard. 





\begin{table}
  \centering
\begin{tabular}{c|cc|cccccccc}
Method & mAP & NDS & Car & Ped & Bus & Barrier & T. C. & Truck & Trailer & Moto.  \\ 
\hline

PointPillars\cite{lang2019pointpillars}& 30.50 & 45.30 & 68.40 & 59.70 & 28.20 & 38.90 & 30.80 & 23.00 & 23.40 & 27.40 \\
InfoFocus\cite{wang2020infofocus}& 39.50 & 39.50 & 77.90 & 63.40 & \textbf{44.80} & 47.80 & 46.50 & 31.40 & 37.30 & 29.00 \\
PointPillars+\cite{vora2020pointpainting}& 40.10 & 55.00 & 76.00 & 64.00 & 32.10 & 56.40 & 45.60 & 31.00 & 36.60 & 34.20 \\ 
\begin{tabular}[c]{@{}c@{}}\textbf{Self-supervised} \\ \textbf{PointPillars}\end{tabular} & \textbf{43.63} & \textbf{56.28} & \textbf{81.00} & \textbf{73.10} & 37.10 & \textbf{58.20} & \textbf{47.80} & \textbf{36.10} & \textbf{41.80} & \textbf{35.40} \\ \hline

CenterPoint\cite{vora2020pointpainting}& 49.54 & 59.64 & 83.40  & 76.10  & 54.20 & 62.40 & 62.40 & 44.40 & \textbf{48.90} & 37.80 \\ 
\begin{tabular}[c]{@{}c@{}}\textbf{Self-supervised} \\ \textbf{CenterPoint}\end{tabular} & \textbf{51.42} & \textbf{60.92} & \textbf{83.80} & \textbf{77.00} &  \textbf{56.80} & \textbf{65.10} & \textbf{63.90} & \textbf{46.30} & 48.50& \textbf{41.10} \\ \hline

\end{tabular}
  \caption{Self-supervised PointPillars and CenterPoint results on nuScenes test set comparing with other PointPillars-based detector and CenterPoint baseline submissions from the nuScenes leaderboard. The  CenterPoint baseline is from \href{https://github.com/open-mmlab/mmdetection3d/tree/v1.0.0.dev0/configs/centerpoint}{mmdetection3d}.}
  \label{tab:nuscenes_test}
\end{table}

\vspace{-1.5cm}

\subsection{Ablation Study}
We conduct two types of ablation studies to further justify the effectiveness of our self-supervised pre-training approach: (i) performance after training with limited annotated data and (ii) performance with and without alternating training strategy. Datasets with 3D annotations are mostly limited for the real-world scenarios due to expense and difficulty of requiring expert knowledge for the annotation process. To show our method's enhancement over the baseline using the self-supervised pre-training, we train our self-supervised 3D detectors and baselines with a percentage of the annotated datasets. 

\begin{table}
  \centering
  \begin{tabular}{cccccccccc}
   Training Data Size  & \multicolumn{3}{c}{1\% } & \multicolumn{3}{c}{5\% } & \multicolumn{3}{c}{20\% }  \\ \hline
    
\multicolumn{1}{c|}{Car AP (IoU=0.7)} & Easy & Mod  & \multicolumn{1}{c|}{Hard} & Easy & Mod  & \multicolumn{1}{c|}{Hard} & Easy & Mod  & Hard  \\ \hline
\multicolumn{1}{c|}{Point-GNN} & 63.34 & 50.92 & \multicolumn{1}{c|}{44.05} & 81.26 & 71.27 & \multicolumn{1}{c|}{65.05} & 88.47 & 77.20 & 74.20 \\ 

\multicolumn{1}{c|}{\textbf{SSL Point-GNN}} & \textbf{66.47} & \textbf{51.42} & \multicolumn{1}{c|}{\textbf{44.63}} & \textbf{84.04} & \textbf{72.69} & \multicolumn{1}{c|}{\textbf{65.93}} & \textbf{88.65} & \textbf{79.52} & \textbf{74.87} \\ \hline

\multicolumn{1}{c|}{Improvement} & \multicolumn{1}{c}{{\color[HTML]{009901} +3.13}} & \multicolumn{1}{c}{{\color[HTML]{009901} +0.50}} & \multicolumn{1}{c|}{{\color[HTML]{009901} +0.58}}  & \multicolumn{1}{c}{{\color[HTML]{009901} +2.78}} & \multicolumn{1}{c}{{\color[HTML]{009901} +1.42}} & \multicolumn{1}{c|}{{\color[HTML]{009901} +0.88}} & \multicolumn{1}{c}{{\color[HTML]{009901} +0.18}} & \multicolumn{1}{c}{{\color[HTML]{009901} +2.32}} & \multicolumn{1}{c}{{\color[HTML]{009901} +0.67}} \\ \hline

\end{tabular}
  \caption{Self-supervised (SSL) Point-GNN trained with a percentage ($1\%$, $5\%$, and $20\%$) of labelled 3D detection data. 3D AP$_{R_{40}}$ results for car class on KITTI val. set.}
  \label{tab:kitti_5perc_data}
\end{table}

\vspace{-0.8cm}

In Table \ref{tab:kitti_5perc_data}, we show the performance of self-supervised Point-GNN and the baseline trained with $1\%$, $5\%$, and $20\%$ of the KITTI train split. Our method consistently outperforms the baseline for all difficulty levels on KITTI validation set.  We note that all self-supervised 3D detector ablation results are obtained without alternating training except the alternating training ablation. We conduct the same experiment for PointPillars, CenterPoint, and SSN 3D detectors on nuScenes validation set and report the results in the supplementary material. Similarly, self-supervised 3D detectors outperform their baselines with large margins. Overall, our results suggest that the self-supervised scene flow pre-training can help learn more representative point-wise features in the lack of labelled training data.



In addition, we conduct an ablation study to justify our alternating training strategy. The alternating training enhances the hard difficulty 3D AP with $2.32\%$ increment for Point-GNN. We think that this improvement is due to the repeated motion-awareness of the backbone brought by the first 3D detection fine-tuning. The detailed 3D and BEV AP results for Point-GNN are provided in the supplementary material. Similarly, the alternating training results for PointPillars, CenterPoint, and SSN 3D detectors reported in the supplementary material support our argument. 

\subsection{Comparison with Other Self-supervised Learning Methods}
Our method is the first study that shows the relation between the self-supervised scene flow and 3D detection representations. Our experiments show that the self-supervised scene flow pre-training provides useful point representations for the supervised 3D detection training. In addition, we compare our CenterPoint-based self-supervised scene flow pre-training against other state-of-the-art self-supervised learning methods in Table \ref{tab:ssl-comparison}. Our method performs better than other CenterPoint-based self-supervised methods in the low-data regime on the nuScenes validation set. 

\begin{table}[h]
\centering
\begin{tabular}{c|c|cc|cc}
Approach & Model &\multicolumn{2}{c|}{5\%} & \multicolumn{2}{c}{10\%} \\ \cline{3-6}
& & \multicolumn{1}{c|}{mAP} & NDS & \multicolumn{1}{c|}{mAP} & NDS \\ \hline
PointContrast\cite{xie2020pointcontrast}  & & \multicolumn{1}{c|}{30.79} & 41.57 & \multicolumn{1}{c|}{38.25} & 50.1 \\
GCC3D\cite{liang2021exploring} & CenterPoint\cite{yin2021center} & \multicolumn{1}{c|}{32.75} & 44.2 & \multicolumn{1}{c|}{39.14}    & 50.48                                 \\
\textbf{Ours} & & \multicolumn{1}{c|}{{\color[HTML]{333333} \textbf{36.04}}} & {\color[HTML]{333333} \textbf{48.28}} & \multicolumn{1}{c|}{{\color[HTML]{333333} \textbf{41.29}}} & {\color[HTML]{333333} \textbf{51.35}} \\ \hline
\end{tabular}
\caption{Comparison against other self-supervised learning methods on nuScenes validation set. GCC3D and PointContrast results are taken from \cite{liang2021exploring}. }
\label{tab:ssl-comparison}
\end{table}

\subsection{Sparse Scene Flow Estimations}
\begin{figure*}[t]

  \centering
  \includegraphics[width=0.9\linewidth]{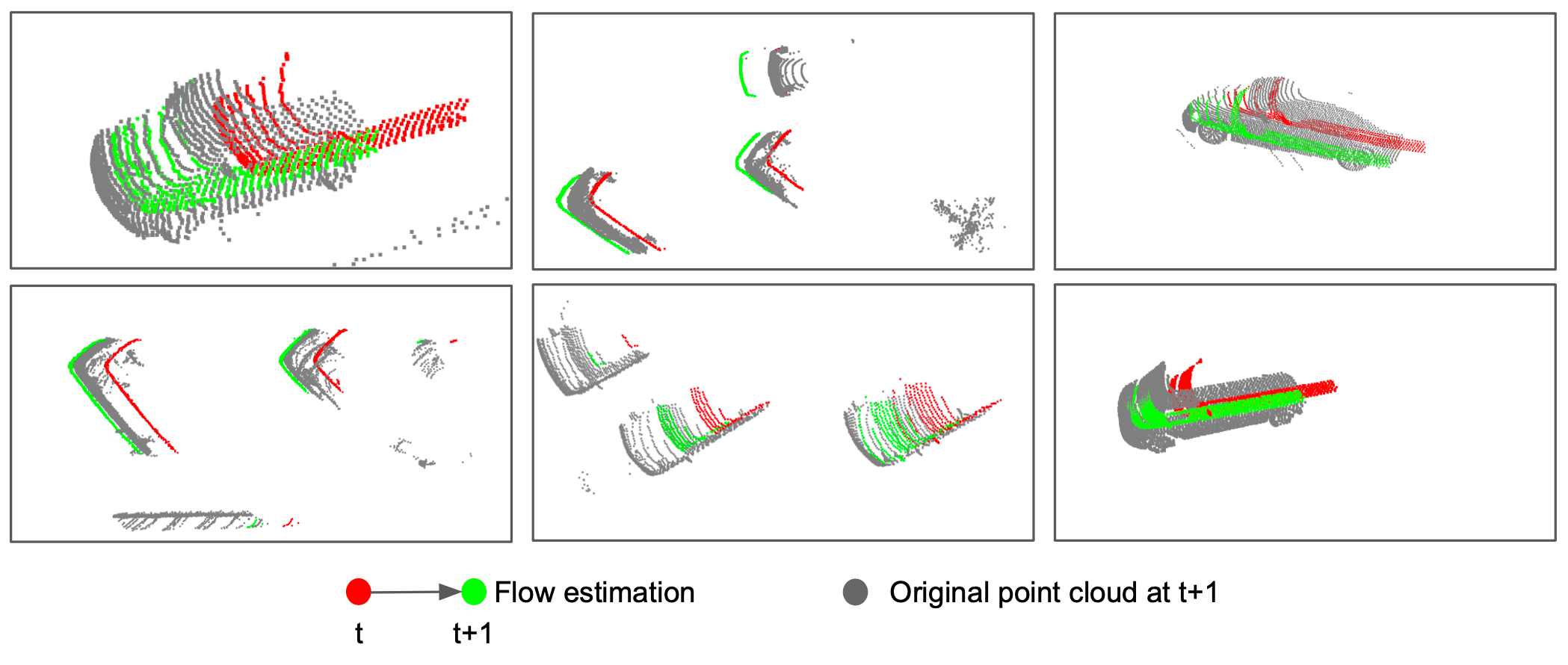}
  \caption{Sparse scene flow estimation on the sampled KITTI lidar points from different frames. Gray points are from the full point clouds at frame $t+1$, red points are sampled points at frame $t$, and green ones are the propagated points to the frame $t+1$ using scene flow estimation.}
  \label{fig:flow}
\end{figure*}

Fig. \ref{fig:flow} shows visualized sparse scene flow estimations on the sampled KITTI lidar point clouds obtained using Point-GNN backbone. Red points are the sparsely-sampled points at frame $t$, which are propagated to the frame $t+1$ using the estimated flow vectors as shown with the green points. The green points closely match the gray points, which are the original point cloud at frame $t+1$. The network is trained with the cycle consistency loss followed by a $100$ epoch fine-tuning on the KITTI Scene Flow Dataset following \cite{mittal2020just}. This suggests that our scene flow network learns useful point features and therefore the point cloud motion patterns, which improves the 3D object detection accuracy. 


\section{Conclusion}
In this study, we propose a self-supervised backbone training approach for 3D object detection. We utilize large unlabelled datasets for self-supervised training of the 3D detection backbone. The scene flow task is used for the self-supervision using the cycle consistency, which helps the backbone learning the point cloud data structure. We show that our approach can improve the detection results of different 3D detectors comparing to their baselines on KITTI and nuScenes datasets. We also show that self-supervised pre-training is especially helpful with the lack of data. Our approach is flexible and can be combined with different point- and voxel-based 3D detectors.



\clearpage
%
%
\bibliographystyle{splncs04}
\bibliography{egbib}
\end{document}